# Context-Aware Adaptive Sampling for Intelligent Data Acquisition Systems Using DQN

Weiqiang Huang
Northeastern University
Boston, USA

Juecen Zhan
Vanderbilt University
Nashville, USA

Yumeng Sun
Rochester Institute of Technology
Rochester, USA

Xu Han
Brown University
Providence, USA

Tai An
University of Rochester
Rochester, USA

Nan Jiang*
Carnegie Mellon University
Pittsburgh, USA

*Abstract- Multi-sensor systems are widely used in the Internet of Things, environmental monitoring, and intelligent manufacturing. However, traditional fixed-frequency sampling strategies often lead to severe data redundancy, high energy consumption, and limited adaptability, failing to meet the dynamic sensing needs of complex environments. To address these issues, this paper proposes a DQN-based multi-sensor adaptive sampling optimization method. By leveraging a reinforcement learning framework to learn the optimal sampling strategy, the method balances data quality, energy consumption, and redundancy. We first model the multi-sensor sampling task as a Markov Decision Process (MDP), then employ a Deep Q-Network to optimize the sampling policy. Experiments on the Intel Lab Data dataset confirm that, compared with fixed-frequency sampling, threshold-triggered sampling, and other reinforcement learning approaches, DQN significantly improves data quality while lowering average energy consumption and redundancy rates. Moreover, in heterogeneous multi-sensor environments, DQN-based adaptive sampling shows enhanced robustness, maintaining superior data collection performance even in the presence of interference factors. These findings demonstrate that DQN-based adaptive sampling can enhance overall data acquisition efficiency in multi-sensor systems, providing a new solution for efficient and intelligent sensing.*

I. INTRODUCTION

With the rapid growth of IoT technologies, numerous distributed sensors are now deployed in key fields such as smart cities, intelligent manufacturing, environmental monitoring, and healthcare [1]. These sensors primarily handle data collection. The data they gather directly affects the accuracy and efficiency of subsequent analysis, decision-making, and system responses. Traditional sensor sampling strategies often rely on fixed frequencies or simple threshold triggers [2]. They lack adaptability to changing environments. In dynamic settings, such static approaches lead to redundant data, missed critical information, and unnecessary resource consumption. They also fail to meet the demands of modern intelligent systems for efficient and accurate data acquisition. Therefore, designing an intelligent strategy that can dynamically adjust sampling behavior based on environmental changes has become a crucial research issue in sensor data collection [3].

In multi-sensor systems, scheduling sampling resources becomes especially complex. Different sensor types have varied energy consumption and information contributions. Potential spatiotemporal redundancy or complementarity also exists among sensors. A uniform static sampling strategy thus struggles to optimize overall performance. Moreover, constraints such as energy, bandwidth, and computing capacity further limit resource availability [4]. Consequently, a method is needed to allocate limited sampling resources and dynamically adjust each sensor's sampling behavior, while still meeting task requirements. Existing rule-based heuristic strategies lack the flexibility and generalization needed for complex, evolving environments. Against this backdrop, reinforcement learning, with its self-learning capabilities, offers a promising new approach for adaptive sampling optimization [5].

Reinforcement learning has rapidly become a prominent intelligent decision-making tool in recent years. It excels at learning optimal policies through interaction with changing environments, showing strong adaptability and long-term optimization capabilities. In particular, Deep Q-Network (DQN) integrates deep neural networks with Q-learning, enabling reinforcement learning to function effectively in high-dimensional state spaces [6-9]. In multi-sensor sampling optimization, DQN can perceive dynamic system states such as environmental changes, task demands, and remaining sensor energy. Drawing on historical experience, it learns a dynamic decision strategy that optimally controls sampling frequency and sensor activation modes. Compared with traditional methods, DQN provides stronger generalization and intelligence, allowing more efficient resource allocation and higher data collection efficiency in complex environments.

From an application standpoint, DQN-based adaptive sampling strategies can significantly reduce system energy consumption and communication overhead. They can also enhance the detection rate and response speed for critical events, delivering substantial practical benefits. For instance, environmental monitoring sensors should sample more frequently when pollution levels change sharply, yet conserve energy during stable periods [10]. In remote health monitoring, wearable devices must adjust ECG or temperature sampling density based on shifts in physiological indicators, achieving the goal of "more precision when it matters, and lower usage when it doesn't." Such applications demand highly responsive and adaptive sampling, which is precisely where reinforcement learning excels [11].

In summary, to address the limitations of traditional sensor sampling strategies in complex environments, this paper proposes a multi-sensor adaptive sampling optimization method based on DQN. By incorporating reinforcement learning, the system gains the ability to autonomously learn sampling policies in different environmental states, thereby improving overall data collection efficiency and resource utilization. This study not only offers theoretical innovation but also has broad engineering applicability. It provides a new technological pathway to building high-efficiency, intelligent sensing systems for the IoT.

## II. METHOD

To implement an efficient adaptive sampling strategy for multi-sensor systems, we model the sampling optimization task as a Markov Decision Process (MDP) within a reinforcement learning framework. Building on previous research that has successfully applied dynamic decision-making models to complex and context-aware environments [12-14], we structure the adaptive sampling problem to capture the sequential interactions between the agent (sampling policy) and the environment (sensor system). In this formulation, the agent observes the system's current state, selects a sampling action, and receives feedback through a reward signal designed to balance data quality, energy efficiency, and redundancy. This MDP formulation provides a principled basis for training the Deep Q-Network (DQN) used in our approach. The structure of the Markov Decision Process is illustrated in Figure 1:

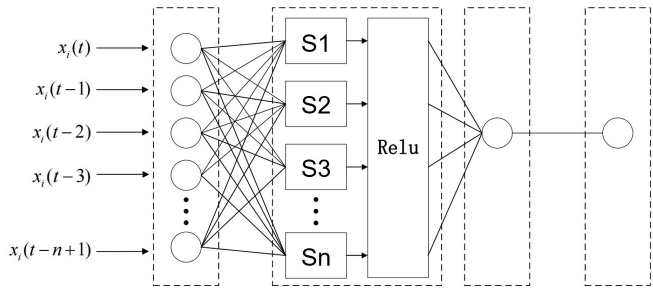

Figure 1. General algorithm flow of Markov process

Figure 1 illustrates the general algorithmic flow of the Markov Decision Process (MDP) within the deep reinforcement learning framework. This structure serves as the foundation for the Deep Q-Network (DQN) applied in our method. Drawing on previous research that emphasizes the importance of temporal feature extraction and sequential modeling in dynamic systems [15-17], DQN processes historical sequences of states as input, extracts key features through a neural network, and outputs Q-values corresponding to each state-action pair. These Q-values are used to evaluate and guide action selection. As depicted in Figure 1, the core mechanism involves feeding the states across multiple time steps into the network, mapping them into a hidden representation, and calculating the Q-values, thereby enabling action selection and policy optimization based on the Markov decision process.

This paper constructs the problem as a five-tuple $(S, A, P, R, \gamma)$. Among them, S represents the environmental state space, which covers the characteristic information such as the current observation value, remaining power, and historical sampling records of each sensor; A represents the action space, which is defined as the sampling activation decision of all sensors at each time step (such as whether to sample, sampling frequency level, etc.); $P(s'|s,a)$ is the state transition probability, which represents the probability of transitioning from state s to state $s'$ after executing action a; $R(s,a)$ is the immediate reward function, which is used to measure the data quality, energy consumption cost and task relevance brought by the current action; $\gamma \in [0,1]$ is the discount factor, which is used to balance long-term and short-term benefits.

On this basis, this paper uses the deep Q network (DQN) algorithm to approximate the optimal strategy. DQN uses a parameterized deep neural network $Q(s,a;\theta)$ to approximate the action value function $Q^*(s,a)$. The goal is to learn an action selection strategy that can maximize the long-term cumulative reward in each state. The objective function is defined as minimizing the residual of the Bellman equation, that is:

$$L(\theta) = E_{(s,a,r,s') \sim D}[(y - Q(s,a;\theta))^2]$$

Further, define the target Q value as:

$$y = r + \gamma \max_{a'} Q(s',a';\theta^-)$$

Among them, $\theta^-$ represents the parameters of the target network, which are regularly copied from the current network, and D is the experience replay pool, which stores historical interaction samples to improve training stability and data efficiency. By continuously interacting with the environment, sampling samples, and updating parameter $\theta$, a reinforcement learning model with good sampling strategies in different environmental states is finally obtained.

To achieve efficient sampling control in multi-sensor systems, the reward function designed in this paper comprehensively incorporates three key factors: information contribution, energy consumption cost, and sampling redundancy. Inspired by prior studies that emphasized the importance of multi-objective optimization and adaptive decision-making in dynamic environments [18-20], the reward function balances the need to maximize valuable information acquisition while minimizing energy expenditure and avoiding redundant data collection. This careful design ensures that the agent can learn a sampling strategy that dynamically adapts to

environmental changes while maintaining high efficiency. The reward function is specifically expressed as:

$$R(s,a) = \lambda_1 \cdot I(a) - \lambda_2 \cdot C(a) - \lambda_3 \cdot D(a)$$

Among them, $I(a)$ represents the information gain or task performance improvement brought by the current action, $C(a)$ is the energy consumption or communication cost brought by sampling, $D(a)$ represents the data duplication caused by redundant sampling between multiple sensors, and $\lambda_1, \lambda_2, \lambda_3$ is the weight coefficient, which controls the importance of each item in the total reward. Through this reward structure, DQN can automatically balance data quality and resource overhead during training, thereby learning the optimal multi-sensor collaborative sampling strategy.

In order to further improve learning stability and strategy convergence speed, a soft update mechanism of the target strategy is introduced, and its parameter update process is expressed as:

$$\theta^- \leftarrow \tau\theta + (1-\tau)\theta^-$$

Among them, $\theta$ is the parameter of the current Q network, $\theta^-$ is the parameter of the target network, and $\tau \in (0,1)$ is the soft update factor, which ensures that the target network follows the changes of the main network in a smooth manner [21], thereby reducing the instability during the training process. In addition, in the decision execution stage, in order to achieve a balance between exploration and utilization, the ε-greedy strategy is used to sample actions, and its expression is:

$$at = \begin{cases} \arg\max aQ(st,a;\theta), & \text{with probability } 1-\varepsilon \\ random \text{ action from A}, & \text{with probability } \varepsilon \end{cases}$$

Among them, $\varepsilon$ controls the exploration intensity and gradually decays as the training progresses, so that the model conducts extensive exploration in the early stage and gradually tends to use the learned strategy in the later stage. The above mechanisms jointly build a stable and efficient reinforcement learning optimization framework, so that the learned strategy has good generalization ability and robustness in dynamic and complex multi-sensor environments.

III. EXPERIMENT

A. Datasets

In this study, we selected the Intel Lab Data as the experimental dataset. It was collected by 54 Mica2Dot wireless sensor nodes deployed by Intel Berkeley Research Lab in 2004. These nodes were located at various positions in a laboratory and continuously measured multiple environmental variables, including temperature, humidity, voltage, and light intensity. Each node sampled data at a fixed frequency and transmitted it wirelessly to a central node, forming a time-series dataset from multiple sensors. The total collection period exceeded one month, yielding more than 2.8 million valid data points. The dataset features high dimensionality, heterogeneous sources, and pronounced dynamic variations.

This dataset inherently exhibits typical multi-sensor scenarios. It includes both spatial distribution differences and temporal data changes, making it well-suited as a benchmark to verify the effectiveness of adaptive sampling strategies. In real-world deployments, sensors often operate at fixed sampling frequencies, which can result in excessive redundant data or missed key changes. Thus, how to dynamically decide whether to sample and at what frequency—based on a node's historical observations and the current environment—becomes a critical issue for improving system efficiency and responsiveness.

During the experiments, we treated the sensor nodes as multiple agents in a reinforcement learning environment and modeled the sampling behavior as a discrete action space. The system state was composed of each node's historical sampling values and their changing trends. The sampling reward was determined by data validity, responsiveness to abrupt environmental changes, and energy consumption. Training and testing on the Intel Lab Data validated that the proposed DQN-based sampling optimization strategy effectively reduces redundant sampling while maintaining task performance, thereby enhancing data collection efficiency and energy control.

B. Experimental Results

In the experimental part, this paper first conducts a data quality and energy consumption comparison experiment under different sampling strategies. The experimental results are shown in Table 1.

Table 1. Data quality and energy consumption comparison experimental results

| Sampling strategy | Average data quality | Average energy consumption (mJ) | Redundancy rate (%) | Critical event detection rate (%) |
|---|---|---|---|---|
| Fixed frequency sampling | 0.72 | 145.3 | 38.6 | 82.1 |
| Random Sampling | 0.65 | 96.7 | 27.4 | 76.4 |
| Threshold Trigger Sampling | 0.78 | 121.5 | 24.9 | 85.6 |
| PPO Adaptive Sampling | 0.87 | 89.2 | 12.3 | 91.8 |
| DQN Adaptive Sampling | 0.83 | 93.6 | 15.1 | 89.3 |

From the results in Table 1, it is clear that different sampling strategies show marked differences in data quality, energy consumption, and event detection capability. Although fixed-frequency sampling achieves relatively high data quality (0.72) and event detection rate (82.1%), it lacks adaptability to environmental changes. As a result, it reaches the highest energy consumption (145.3 mJ) and a high redundancy rate (38.6%). This indicates the collection of a large amount of unnecessary data, leading to resource wastage.

In contrast, the DQN and PPO [22-23] adaptive sampling strategies outperform others in multiple dimensions. PPO achieves the best average data quality (0.87) and event detection rate (91.8%), while maintaining the lowest energy consumption (89.2 mJ) and redundancy rate (12.3%). This finding suggests strong responsiveness to unexpected events and a rational adjustment of sensor frequencies. DQN also demonstrates stable performance, maintaining a high event detection rate (89.3%) and low redundancy (15.1%), with energy consumption controlled at 93.6 mJ. These results confirm that DQN's learned strategy offers solid practicality and effective resource optimization.

Traditional random sampling and threshold-triggered methods consume less energy than fixed-frequency sampling. However, they suffer from poorer data quality and detection rates. Random sampling [24], in particular, experiences large fluctuations in data integrity. This outcome highlights that lacking a clear strategic framework can weaken a system's ability to capture critical information. Overall, adaptive strategies—especially those based on reinforcement learning—demonstrate superior comprehensive performance in multi-sensor environments. They prove the feasibility and advantages of achieving efficient data collection and energy utilization in dynamic scenarios.

Secondly, this paper conducted an experiment to analyze the impact of the reward function weight on system performance, and the experimental results are shown in Figure 2.

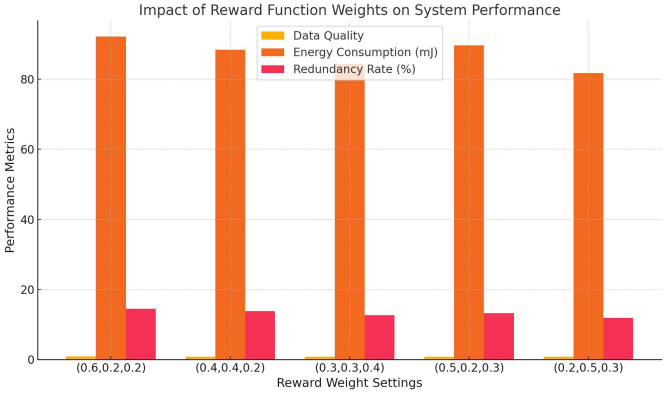

Figure 2. Experimental results of analyzing the impact of reward function weight on system performance

From the experimental results, it is evident that different reward function weight combinations significantly affect the system's performance metrics. First, regarding data quality, when $\lambda_1$ (information gain weight) is relatively large (e.g., (0.6, 0.2, 0.2)), the system achieves the highest average data quality. This indicates that emphasizing information gain can enhance sampling decisions. However, as $\lambda_3$ (redundancy penalty) increases (e.g., (0.3, 0.3, 0.4)), data quality decreases. This suggests that overly strict control of sampling redundancy can lead to loss of critical data, ultimately affecting the overall information quality of the system.

For energy consumption, the results show that when $\lambda_2$ (energy consumption weight) is high, energy use drops noticeably. For example, under the (0.2, 0.5, 0.3) setting, the system consumes the least energy, but data quality and key event detection both decline. This finding suggests that focusing too heavily on energy optimization can cause overly conservative sampling, undermining system effectiveness.

Meanwhile, with a (0.4, 0.4, 0.2) weight combination, energy consumption remains low and data quality remains relatively high, illustrating that certain configurations can balance data quality with energy efficiency.

Changes in redundancy rate show that increasing $\lambda_3$ (redundancy weight) can effectively reduce redundant data, but if redundancy is over-controlled, data quality and environmental adaptability may suffer. Overall, moderate balancing of reward weights for data quality, energy consumption, and redundancy—such as (0.4, 0.4, 0.2) and (0.5, 0.2, 0.3)—can ensure high data quality while lowering energy use and data redundancy. This leads to a more optimal adaptive sampling strategy.

Next, this paper presents a robustness evaluation experiment in a multi-sensor heterogeneous environment, and the experimental results are shown in Figure 3.

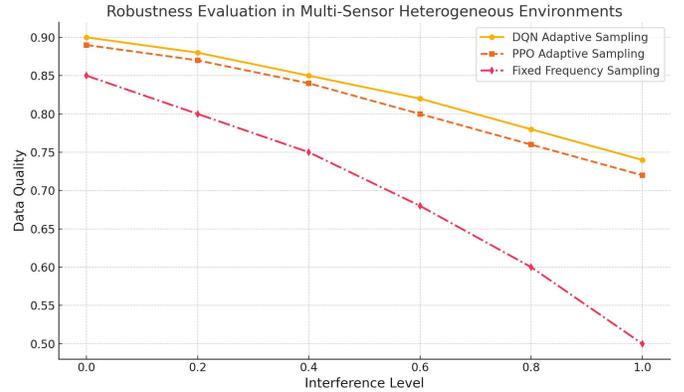

Figure 3. Robustness evaluation experiment in multi-sensor heterogeneous environment

From the experimental results, we see that in a multi-sensor heterogeneous environment, as interference increases, the data quality of all sampling strategies declines. However, their robustness differs. Fixed-frequency sampling experiences the sharpest drop in performance. Under high interference (interference level 1.0), its data quality falls to around 0.5, indicating that it lacks adaptability to environmental changes and is easily affected by external noise. By contrast, PPO and DQN adaptive sampling strategies show smaller declines, suggesting that reinforcement learning methods can adjust sampling decisions dynamically and maintain high data quality.

A closer comparison of DQN and PPO reveals that DQN maintains slightly higher data quality across the entire range of interference levels, especially in the lower range (0.0 to 0.6). This may be because DQN focuses more on long-term returns during training, allowing it to devise more stable sampling strategies in changing environments. At higher interference levels (0.8 to 1.0), the performance gap between the two narrows. This outcome suggests that reinforcement learning strategies overall exhibit strong robustness, adapting effectively to complex sensor environments and mitigating the impact of external interference on data collection.

## IV. Conclusion

This study proposes a multi-sensor adaptive sampling optimization method based on DQN and verifies its performance in various experimental scenarios. The results show that, compared with traditional fixed-frequency sampling, DQN adaptive sampling maintains high data quality while reducing energy consumption and sampling redundancy. It also displays strong adaptability and robustness under different reward weights and heterogeneous multi-sensor environments. Further comparisons with PPO and other reinforcement learning methods confirm DQN's stability and long-term optimization capability in dynamic settings. This finding offers a promising solution for efficient data collection in intelligent sensor systems.

Despite the progress made, there remain several avenues for further improvement. First, DQN's decision-making efficiency and training stability are still limited by the high-dimensional state space. Future research could incorporate attention mechanisms or self-supervised learning to enhance the sampling strategy's generalization. In addition, most experiments were conducted using simulated data. Real-world sensor systems often involve more complex environmental dynamics and noise. Consequently, how to optimize sampling strategies and improve adaptability in actual deployments remains a topic that warrants deeper investigation. Future studies can broaden the application of this method to fields such as intelligent transportation, industrial IoT, environmental monitoring, and healthcare. In these domains, reinforcement learning can further optimize multi-sensor data collection. Combining this approach with distributed reinforcement learning frameworks could also enable sensor nodes to learn cooperatively, thereby boosting overall system performance. As IoT technologies continue to advance, adaptive sampling will play an increasingly important role in efficient sensing, low-power operation, and intelligent data processing. This progress will provide strong support for building more intelligent sensing systems.